**Forecasting Battery Electric Vehicle Charging Behavior:**
**A Deep Learning Approach Equipped with Micro-Clustering and SMOTE Techniques**

**Hanif Tayarani**
University of California, Davis, Institute of Transportation Studies
1590 Tilia St, Davis, California, United States, 95616
Email: htayarani@ucdavis.edu
https://orcid.org/0000-0002-2414-504X

**Trisha V. Ramadoss**
University of California, Davis, Institute of Transportation Studies
1590 Tilia St, Davis, California, United States, 95616
Email: tvramadoss@ucdavis.edu
https://orcid.org/0000-0001-5863-5840

**Vaishnavi Karanam**
University of California, Davis, Institute of Transportation Studies
1590 Tilia St, Davis, California, United States, 95616
Email: vckaranam@ucdavis.edu
https://orcid.org/0000-0002-4647-5250

**Gil Tal**
University of California, Davis, Institute of Transportation Studies
1590 Tilia St, Davis, California, United States, 95616
Email: gtal@ucdavis.edu
https://orcid.org/0000-0001-7843-3664

**Christopher Nitta**
University of California, Davis, Department of Computer Science
3015 Kemper Hall, Davis, California, United States, 95616
Email: cjnitta@ucdavis.edu
https://orcid.org/0000-0003-1531-2771

Word Count: 6476 words + 4 tables (250 words per table) = 7476 words

Submission Date: 07/26/2022

**ABSTRACT**

Energy systems, climate change, and public health are among the primary reasons for moving toward electrification in transportation. Transportation electrification is being promoted worldwide to reduce emissions. As a result, many automakers will soon start making only battery electric vehicles (BEV). BEV adoption rates are rising in California, mainly due to climate change and air pollution concerns. While great for climate and pollution goals, improperly managed BEV charging can lead to insufficient charging infrastructure and power outages. This study develops a novel Micro-Clustering Deep Neural Network (MC-DNN), an artificial neural network algorithm that is highly effective at learning BEV's trip and charging data to forecast BEV charging events – information that is essential for electricity load aggregators and utility managers to provide charging stations and electricity capacity effectively. The MC-DNN is configured using a robust dataset of trips and charges that occurred in California between 2015 and 2020 from 132 BEVs, spanning 5 BEV models for a total of 1,570,167 vehicle miles traveled. The numerical findings revealed that the proposed MC-DNN is more effective than benchmark approaches in this field, such as support vector machine, k-nearest neighbors, decision tree, and other neural network-based models in predicting the charging events.

**Keywords:** Battery Electric Vehicle, Charging Level, Deep Learning, Imbalanced Classification, Transportation

## I. INTRODUCTION

### A. Background and Motivation

The transportation sector has the third-highest energy consumption (*1*) among all industries and is responsible for a significant share of greenhouse gas emissions and air pollution (*2*). Transportation electrification is motivated by concerns over the detrimental effects of transportation on energy systems, climate change, and public health. Thus, electrifying transportation is being pursued around the globe, to curb emissions, and soon there will be many car manufacturers producing only electric vehicles. The market penetration of battery electric vehicles (BEVs) in the United States reached 2.5% in 2022 and is anticipated to reach 50% by 2050. However, California, a leader in Zero Emission Vehicle policy, reached 10% in 2022 and aims for 100% zero emission vehicles including BEVs plug-in hybrids and fuel cell vehicles in 2045 (*3*). Thus, BEVs are projected to grow rapidly. However, this fast expansion will create new challenges in designing and operating charging infrastructure.

It's essential to get an accurate estimate of how these new vehicles will be charged so that problems with the power grid, like unstable voltage and power loss, can be avoided. This will also help prepare the infrastructure for these new vehicles. It is also essential to estimate the quantity of charging infrastructure needed by charging level to prepare the electric grid for the emergence of BEVs and how much money needs to be invested in charging infrastructure. BEVs can be charged in three different charging categories: Level 1, Level 2, and DC-Fast, each having a unique charging speed and impact on the grid (*4*). Trip distance, arrival-departure time, and charging behavior are key features of individual BEV trips that play a vital role in shaping electricity demand from BEVs. Historically, travel demand models were primarily used to forecast daily trip and activity patterns (*5*). However, these models are not suitable for modeling BEV behavior because a higher degree of accuracy is needed to predict and manage BEV electric loads.

### B. Related Work

Electric utilities and decision-makers need to accurately estimate the number of charging stations to develop charging infrastructure (*6*). This is necessary to supply the electricity that the BEVs will demand. For this strategy to be effectively enforced, electricity load aggregators need accurate information on BEVs' charging patterns, specifically the likelihood of being charged upon reaching destinations. As the penetration of BEVs in the Californian transportation fleet grows, we are confronted with a big data dilemma in estimating the charging demand required of various types of BEV fleets with different charging patterns (*4*). For optimum planning and operation of BEVs, the challenge of big data in transportation systems must be adequately addressed (*7*).

In the current research, there are three main ways to look at BEV charging demand: deterministic, scenario-based, and data-driven.

In the first approach, a predefined demand pattern is considered for the vehicles. For instance, the authors in (*8*) studied a battery swap station, and a fixed state of charge (SOC) was considered for all the vehicles at the station. Likewise, Cui et al. (*9*) studied the charging station planning problem and considered a fixed rate of BEV demand. These deterministic approaches, due to their basic assumptions about vehicles' charging patterns, could lead to under or over-estimations in their output.

Existing scenario-based studies have mostly oversimplified BEV charging patterns by mapping the travel patterns of internal combustion engines, which may have very different travel and charging behaviors than BEVs because they don't need to be charged. Furthermore, most existing approaches do not consider the charge level when predicting charging demand, which is the crucial characteristic of the charging pattern. Predicting BEV charging events aids infrastructure preparedness for rising demand and helps power system operators optimize generation and prevent utility grid blackouts. These approaches, which use stochastic or probabilistic methods, can be divided into four groups.

The first method uses Monte Carlo simulations to generate a variety of behavioral scenarios (*10*–*12*). The method, however, usually assumes a normal distribution for trip parameters, which ultimately reduces its accuracy (*13, 14*). Furthermore, the problem's dimensions will grow significantly when the number of BEVs increases. Due to the high cost of computing, the algorithm is not helpful for real-world case studies that involve complex operations and planning. Even though many BEV characteristics, like arrival-departure time, destination location, and trip distance, affect each other, these earlier models treated them as separate variables instead of considering how they affect each other. Moreover, since these methods are rooted in generated, artificial behavioral scenarios, they are hindered by computational limitations and may not accurately reflect real-life behavior.

The next approach uses Markov chain theory to predict BEV charging load demand, which, similar to Monte Carlo simulations, neglects the correlation between the travel parameters and has a high computational cost due to the high dimension of the state matrix of the Markov chain (*15*, *16*). Furthermore, the grouping structure of Markov methods in forecasting charging events causes computational limitations that affect the accuracy of their results under high BEV market share.

Another approach in this field is queuing theory. For example, in this method, a homogeneous Poisson model can be modeled by forecasting BEVs' arrival and departure times as a proxy of their travel behavior instead of actual trip length. Furthermore, based on trip characteristics, a charging event will be predicted. Despite the recent progress by considering a non-homogeneous Poisson model that includes the state of charge in batteries to produce more realistic results, this method still doesn't consider how different travel characteristics are related, which makes the prediction of charging events less accurate (*17*).

Finally, the trip chain concept is built on a transportation-based approach to estimate the travel behavior of BEVs. For instance, Wang et al.(*18*) developed a Naïve Bayes-based model that temporally and spatially couples the trip chains to include the departure, arrival, and trip distance using the National Household Travel Survey (NTHS) data (*19*). However, given the source of information, the trip chain method is more suited for internal combustion engine vehicles with different travel behaviors than BEVs. Furthermore, some studies consider simplifying assumptions in modeling BEV charging demand, such as controlled charging over a given time or controlled travel patterns leaving and returning home in the a.m. and p.m. peak hours (*20*). However, employing similar travel patterns as conventional vehicles (*21*) and ignoring the travel distance (*22*) all introduce errors in forecasting BEV travel behavior and, consequently, estimating their charging events.

To exceed the limitations of traditional modeling methods, conventional machine learning as a data-driven approach has been introduced to predict the charging demand (*23*). These models are important to accurately model charging events since the tempo-spatial patterns of BEVs' daily activity introduce complexities and uncertainties in their charging events that are undetectable by conventional methods. Machine learning-based methods range from support vector machine (SVM), random forest (RF), and decision tree (DT) to artificial neural network (ANN) methods (*24*). However, the low classification capacity, lack of strong theoretical background, and not being perfectly accurate are all reasons to move toward using deep learning methods. Indeed, these conventional methods do not have high feature extraction abilities in large-dimension data sets (*25*).

Deep Neural Networks (DNNs) are introduced to overcome the shortcomings of conventional machine learning methods in handling high uncertainties (*26*). DNNs are a type of ANN that can predict dynamic behaviors with higher accuracy. Because of their ability to extract key information, deep architecture is well suited for classification problems. DNNs can process simple non-linear relationships in datasets, making them the ideal candidates for processing our unique trip and charging dataset.

 To improve the accuracy of the DNN, recently, classification approaches as an impressive complement are applied to the DNN methods as a pre-processing technique. By classifying raw data into various predefined clusters and using the clustered output as a new feature, higher accuracy can be reached (*27*). This procedure helps to handle the high uncertainty of data and produce more accurate results (*25*). Despite the efforts in predicting charging behavior by artificial intelligence researchers by developing machine learning algorithms and transportation planners by developing data extraction methods from individual activity, a more accurate method to predict charging behavior, particularly charging level, is still required. This study introduces an innovative micro-clustering deep learning-based approach to forecast BEVs' charging events to fill the knowledge gap created by inaccurate charging event forecasting. The proposed method aims to improve the weaknesses of the statistical and conventional machine learning methods used in previous work. DNN approaches are promising tools for addressing existing challenges in estimating BEV charging behavior by applying big data techniques that consider their capabilities to handle stochasticity, achieve higher accuracy in comparing statistical methods, and be trainable.

### C. Paper Contributions

This study aims to design a DNN equipped with micro clustering and SMOTE technique (*28*) to learn how to classify BEV charging events at the destinations based on their level.

Knowing the charging events and their levels at the end of trips can help operators estimate the number of charging stations required for a new BEV fleet (*29*). Our research introduces a novel deep learning-based method equipped with clustering and SMOTE technique to infer charging events from simply obtainable trip characteristics such as departure and arrival time and SOC. Furthermore, we train our deep learning method with unique fully electric vehicle trip datasets to reveal the logic behind the individuals' activity participation. Concurrently, the proposed method makes charging behavior prediction more accurate and ultimately helps the utility grid to develop the charging stations and could be utilized by power system planners to improve forecasting the BEV charging demand significantly. The main contributions of this paper can be categorized as follows:

- Employing robust unsupervised-supervised clustering tasks to classify the charging events based on charging level.
- The training model is based on real-world charging data from 132 BEVs—five different models—collected over the course of a year.
- Three charge levels, including Level-1, Level-2, and DC-Fast, are considered for the plugged-in event.

The proposed clustering technique can handle the high uncertainty and intermittent behavior of charging patterns. The numerical results verify the robustness of the proposed method.

### D. Paper Organization

The rest of this paper is organized as follows. Section II gives a brief description of the dataset and a comprehensive description of the proposed method in this paper. In Section III, numerical results are presented in detail. An experimental comparison between the benchmark and proposed methods is discussed in Section IV. Finally, Section V concludes this paper.

## II. DATA AND METHODS

### A. Data Description

The dataset used to create the classifiers in this research is a subset of the Advanced Plug-in Electric Vehicle (PEV) Driving and Charging Behavior Project, a five-year (2015–2020) California-wide study aimed at understanding the driving and charging behavior of plug-in electric vehicles (PEVs). This research project collected information from 400 households and 800 cars, including 400 PEVs. A subset of households with at least one BEV was chosen for this study, and each BEV had a data logger installed for around a year. On a second-by-second basis, the logger recorded important driving and charging characteristics such as speed and GPS coordinates. **Table 1** summarizes the loggers' data collected on EV charging and driving. The summaries include vehicle data spanning approximately one year for each logged car, collected over the five-year course of this study (*30*).

**TABLE 1: BEV dataset (*30*)**

|  | EV Type | Number of Vehicles | Number of Trips | Total Miles Traveled | Number of Charging | Total kWh Charged |
|---|---|---|---|---|---|---|
| *Battery Electric Vehicles* | Nissan Leaf-24 | 29 | 34,061 | 262,210 | 8,707 | 57,638 |
|  | Nissan Leaf-30 | 28 | 33,435 | 267,335 | 6,744 | 62,804 |
|  | Chevrolet Bolt 66 | 27 | 39,479 | 381,032 | 8,351 | 100,535 |
|  | Tesla Model S-60_80 | 23 | 21,057 | 374,908 | 6,737 | 139,777 |
|  | Tesla Model S-80_100 | 25 | 20,032 | 284,682 | 4,902 | 106,710 |
|  | All BEVs | 132 | 148,064 | 1,570,167 | 35,441 | 467,464 |

**Figures** 1-a and 1-b show the two most crucial trip characteristics: the trip start time and start SOC. Most trips happen between 8 a.m. and 6 p.m., with an initial SOC above 80 percent.

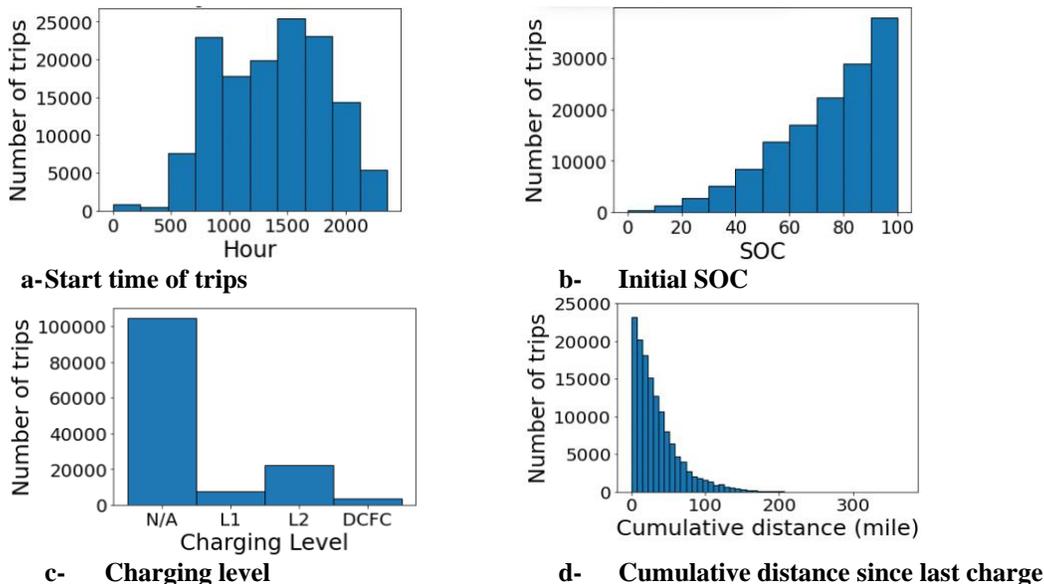

**a- Start time of trips**          **b-     Initial SOC**

**c-     Charging level**          **d-     Cumulative distance since last charge**

**FIGURE 1: Trip and charging dataset summary**

**Figure 1-c** depicts the target variable in this study. This variable shows us that trips are split into None, Level 1, Level 2, and DC-Fast charging events based on their charging behavior at the end of the trip. The large gap between the number of non-charging events and the other charging levels makes the charging prediction problem an imbalanced classification problem, requiring complex tools to handle the imbalanced classes. The histogram of the cumulative distance traveled since the last charge is shown in **Figure 1–d** above.

**B.    Benchmarks Methods**

Five benchmark methods in forecasting tasks are considered in this project to verify the robustness of the proposed method. As a statistical method and a standard technique, the cost-sensitive logistic regression method is also considered (CS-LR). K-Nearest Neighbors (KNN), which is a typical technique in classification, is considered. Additionally, this study looks at neural networks, both shallow (ANN) and deep (DNN) networks, which are often used as benchmark methods in various classification problems. Lastly, support-vector machines (SVMs) are considered as they are often utilized as a robust machine learning technique in classification due to their excellent performance in feature extraction tasks with kernel functions. SVM uses several kernel functions to map input data into a higher-dimensional space. This study looks at polynomial (SVM-Poly) and radial basis (SVM-RBF) kernel functions, which are two predominant types of kernel functions (*31*).

### C. Micro-Clustering Deep Neural Network (MC-DNN)

Two key modules define the proposed method's main structure: micro-clustering (MC) and forecasting modules. This part contains a complete description of these modules.

1- MC Module

Unsupervised and supervised clustering tasks are explained in the proposed MC module. This section describes unsupervised and then supervised clustering tasks.

*Unsupervised Clustering Task:*

First, it's important to remember that each cluster has no labels. Hence, the clustering process must be completed in an unsupervised environment using machine learning techniques like K-means and Mean-Shift Clustering. K-means clustering is faster and more accurate than existing unsupervised clustering methods. The Silhouette index, a strong criterion in this field (*32*), is used to evaluate the robustness of the unsupervised clustering method in this paper. According to the Silhouette index, four is the optimal number of clusters. Also, before clustering, all of the numeric inputs are transferred to a logarithmic scale and normalized between 0 and 1 respectively.

*Supervised Clustering Task:*

At this point, each cluster has a target, and the input data for the charging event forecasting has been assigned to the supervised classification network. DNN, which is utilized in this section, is one of the most popular machine learning techniques for classification tasks, and it performs well in this field (*33*).

2- Forecasting Module

Deep Learning approaches with several representation layers suit the charging prediction problem better than shallow ANNs since they are better at feature extraction. This enables deep learning methods to assess the input data more thoroughly compared to the shallow structure embodied in a single layer ANN. Deep Learning methods best suit solving large dimension problems such as image and speech recognition and diverse forecasting studies. Charging level prediction is also a large dimension study. As a result, we used a deep learning structure to use this powerful feature extraction technique in this study. The proposed deep network is made up of several hidden layers. Because charging event prediction is an imbalanced classification task, after splitting the dataset into train sets and test sets, a SMOTE technique is used to balance the train sets to improve neural network performance (*28*). This is advantageous since a larger number of samples may improve deep learning performance compared to other machine learning approaches. To make the process of making a deep artificial neural network easier to understand, here are the feed-forward equations for a sample neural network with one hidden layer including two neurons (**Figure 2**):

$$net_1^1(s) = (w_1^1(s))^T.X \tag{1}$$

$$net_2^1(s) = (w_2^1(s))^T.X \tag{2}$$

$$O^1(s) = [O_0^1(s), f_1^1(net_1^1(s)), f_2^1(net_2^1(s))]^T \tag{3}$$

$$net_1^2(s) = (w_1^2(s))^T.O^1(s) \tag{4}$$

$$O_1^2(s) = f_1^2(net_1^2(s)) \tag{5}$$

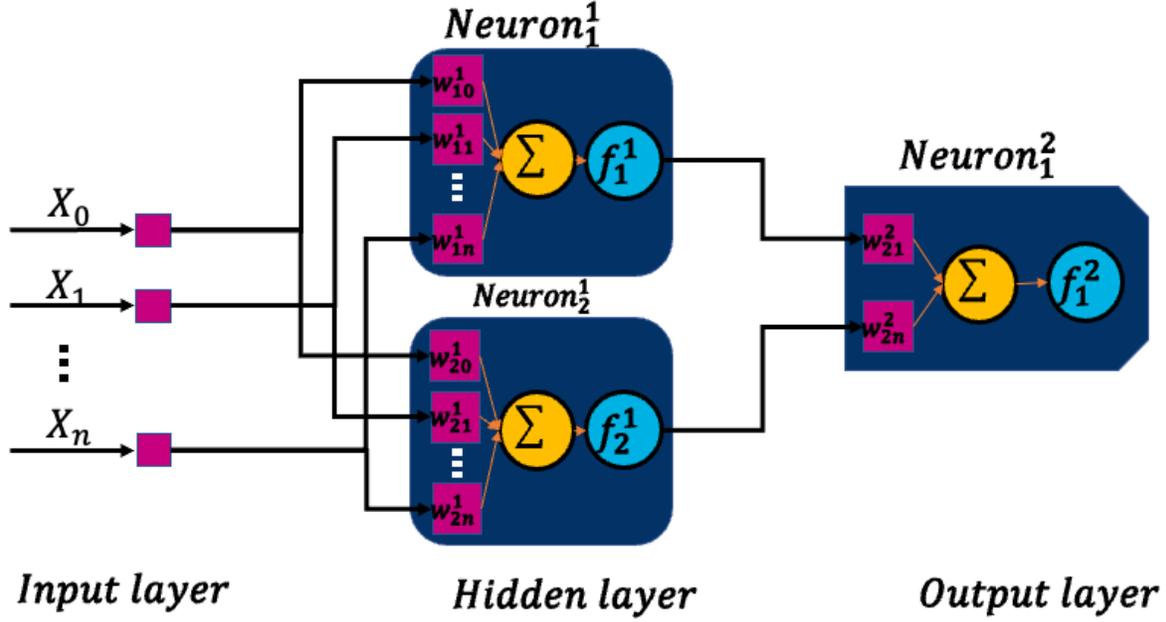

**FIGURE 2: Multilayer Perceptron Neural Network**

The following are the error back propagation equations that define the gradient descent-based learning strategy (*27*).

$$E(s) = -\sum_{c=1}^{M} D(s,c).log(o_1^2(s,c)) \tag{6}$$

$$\nabla \omega_1^2\big(E(s)\big) = \frac{\partial E(s)}{\partial w_1^2(s)} = \frac{\partial E}{\partial net_1^2} \times \frac{\partial net_1^2}{\partial \omega_1^2}(s) \tag{7}$$

$$\nabla \omega_1^1\big(E(s)\big) = \frac{\partial E(s)}{\partial \omega_1^1(s)} = \frac{\partial E}{\partial net_1^2} \times \frac{\partial net_1^2}{\partial o_1^1} \times \frac{\partial o_1^1}{\partial net_1^1} \times \frac{\partial net_1^1}{\partial \omega_1^1}(s) \tag{8}$$

$$\nabla \omega_2^1\big(E(s)\big) = \frac{\partial E(s)}{\partial \omega_2^1(s)} = \frac{\partial E}{\partial net_1^2} \times \frac{\partial net_1^2}{\partial o_1^1} \times \frac{\partial o_1^1}{\partial net_2^1} \times \frac{\partial net_2^1}{\partial \omega_2^1}(s) \tag{9}$$

The weights of networks are then updated as follows:

$$\Delta \omega_j^l(s) = -\eta_{\ \omega} \nabla \omega_j^l\big(E(s)\big) \tag{10}$$

$$\Delta \omega_j^l(s) = \omega_j^l(s+1) - \omega_j^l(s) = -\eta_{\ \omega} \nabla \omega_j^l\big(E(s)\big) \tag{11}$$

$$\omega_j^l(s+1) = \omega_j^l(s) - \eta_{\ \omega} \nabla \omega_j^l(E(s)) \tag{12}$$

**Table 2** defines the variables, parameters, and sets that are used in the equations above:

**TABLE 2: Nomenclature**

| Category | | Definition |
|---|---|---|
| Variables | $D(s)$ | Desired output in iteration $s$ |
| | $D(s,c)$ | Desired output in iteration $s$ for class $c$ |
| | $E(s)$ | Total Sum of error in iteration $s$ |
| | $E$ | Total sum of error |
| | $f_j^l(s)$ | Activation function for neuron j in layer $l$ in iteration $s$ |
| | $net_j^l(s)$ | Activation function input for neuron j in layer $l$ in iteration $s$ |
| | $O_j^l(s)$ | Output of neuron $j$ in layer $l$ in iteration $s$ |
| | $O_j^l(s,c)$ | Output of neuron $j$ in layer $l$ in iteration $s$ for class $c$ |
| | $\omega_j^l(s)$ | Weight vector for neuron $j$ in layer $l$ in iteration $s$ |
| | $\omega_{ij}^l(s)$ | Weight vector between sample $i$ of the input layer and neuron j in hidden layer $l$ in iteration $s$ |
| | $X$ | Input data vector |
| Parameters | $\eta_\omega$ | Training coefficient for weights |
| | $M$ | Total number of classes |
| | $n_0$ | Total number of input data components |
| index | $c$ | Index of class |
| | $i$ | Index of input data |
| | $j$ | Index of hidden layer sample |
| | $s$ | Index of iteration number |
| | $l$ | Index layer number |

Deep networks are excellent tools for forecasting phenomena with high-intermittent behavior, but they have limitations due to many hidden layers, such as instability and overfitting. This research used three commonly used strategies to handle these problems: Mini-Batch Gradient Descent, dropout, and L2 regularization (*34*, *35*).

We train the DNN-based algorithms using the stochastic gradient descent method, which is substantially faster than gradient descent because of its high frequency of updating the training parameters. Instead of going through the whole set of training data, the stochastic gradient descent method iterates over a few randomly chosen training samples to find the optimal solution. Thus, the stochastic gradient descent approach is less likely to trap into shallow local minimum solutions than the gradient descent method, resulting in a more accurate forecast. Dropout and L2 regularization techniques avoid overfitting. In dropout, some neurons are left off during the training procedure, which allows some neurons to be independent of others. L2 regularization penalizes sharp changes in neurons, which aids in avoiding local minimum points (*34*). The entire flowchart of the proposed algorithm, which is created by connecting the MC and forecasting modules, is seen in **Figure 3**. The MC task, as shown in **Figure 3**, is only performed during the training procedure. As a result, the forecasting procedure's computation burden will be unaffected.

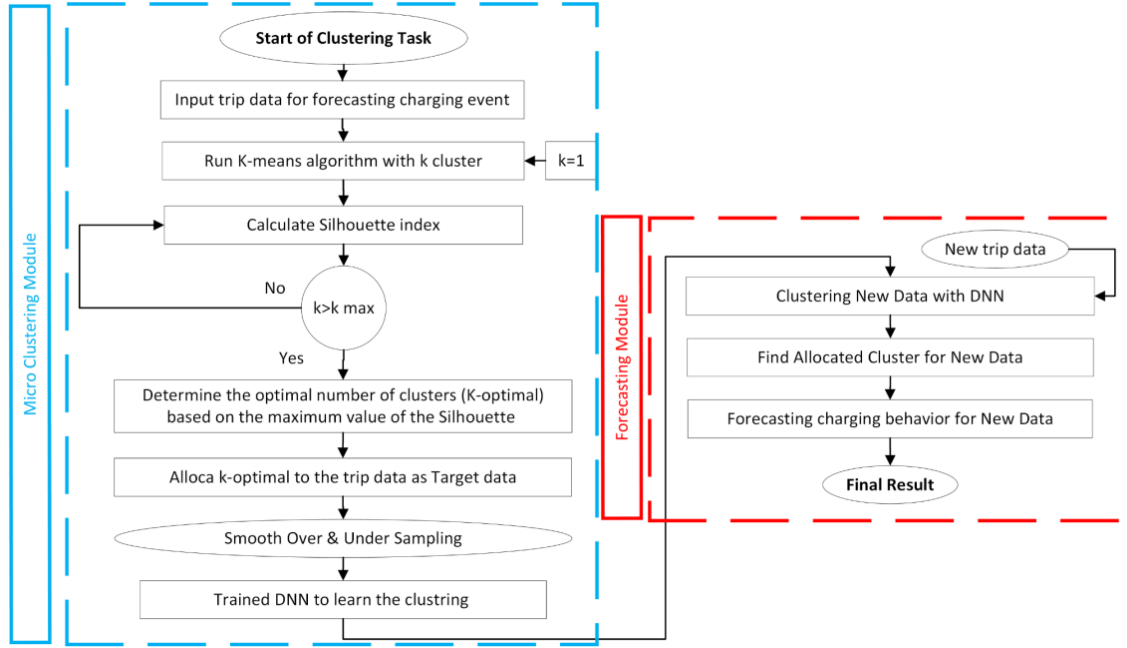

**FIGURE 3: The MC-DNN method's overall flowchart**.

## III. RESULTS

This section first discusses some well-known error criteria that can be used to assess the quality of a forecasting method. Then, the proposed method is compared to benchmark methods that are already in use. In this paper, five error criteria are utilized to assess the robustness of different techniques: accuracy, precision, recall, F-measure, and G-mean. (**Table 3**)

**TABLE 3: Error criteria (*36*)**

| Measures | Formula |
|---|---|
| Accuracy | $\dfrac{TP + TN}{TP + TN + FP + FN}$ |
| Recall | $\dfrac{TP}{TP + FN}$ |
| Precision | $\dfrac{TP}{TP + FP}$ |
| F-Measure | $\dfrac{(\beta^2 + 1) * Recall * Precision}{Recall + \beta^2 * Precision}, \beta > 1$ |
| G-Mean1 | $\sqrt{Recall * Precision}$ |

|  | Predicted class | |
|---|---|---|
|  | P | N |
| Actual Class P | True Positives (TP) | False Negatives (FN) |
| Actual Class N | False Positives (FP) | True Negatives (TN) |

### A. Micro-Clustering Result:

The proposed method starts with the micro clustering task, which is done in unsupervised space by the K-means classifier and in supervised space by the DNN classifier. **Figure 4** depicts the optimal number of input data clusters for the BEV dataset, as indicated by the Silhouette index (the highest value). Due to the complexity of the BEV charging dataset, the vehicle charging and trip behaviors vary in ways like start time, SOC start, distance, destination, etc. In this way, the model needs a new feature, which is extracted by the clustering task, to help the model predict the charging behavior more accurately. In this paper, the maximum number of clusters ($K_{max}$) is assumed to be 7 in unsupervised clustering to make sure the clustering results work well, and bigger ones do not change the clustering results and are not needed.

Silhouette analysis is capable of estimating the separation distance between the generated clusters. The silhouette plot visualizes how each point in one cluster is close to points in another, thus giving a visual way to find an optimal number of clusters. When the silhouette index is close to 0, it means that the sample is close to the line that divides two neighboring clusters. A negative index means that these samples might have been put in the wrong cluster. And finally, an index close to 1 means that the sample is very far from the neighboring clusters. For more information please see (*37*). As depicted in **Figure 4**, the optimal Silhouette index values for start time, SOC start, distance, and destination are achieved on 4 clusters.

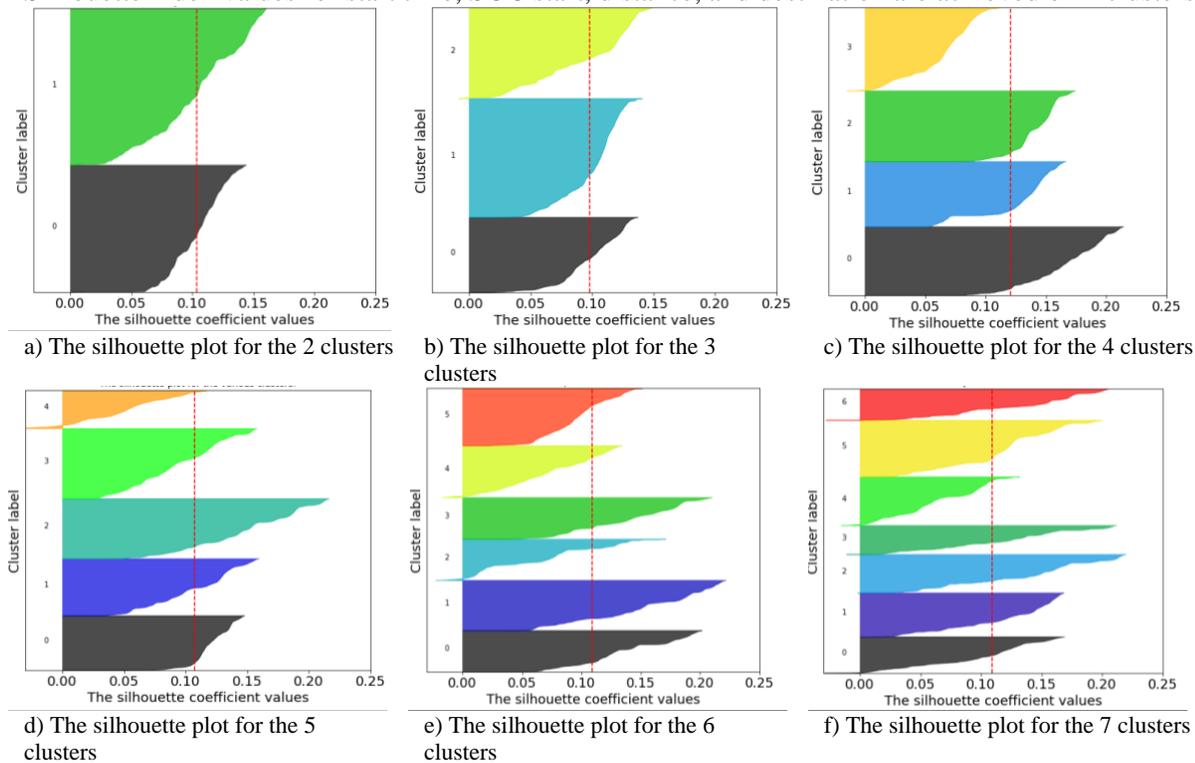

a) The silhouette plot for the 2 clusters

b) The silhouette plot for the 3 clusters

c) The silhouette plot for the 4 clusters

d) The silhouette plot for the 5 clusters

e) The silhouette plot for the 6 clusters

f) The silhouette plot for the 7 clusters

**FIGURE 4: Silhouette analysis for KMeans clustering on the BEV dataset with varying cluster numbers**

Based on the DNN classifier, new features are assigned to the test set. In the supervised clustering problem, the labels are defined as the centroids of each cluster. The confusion matrix illustrates the result of clustering 11,712 randomly selected trips. (See **Figure 5**)

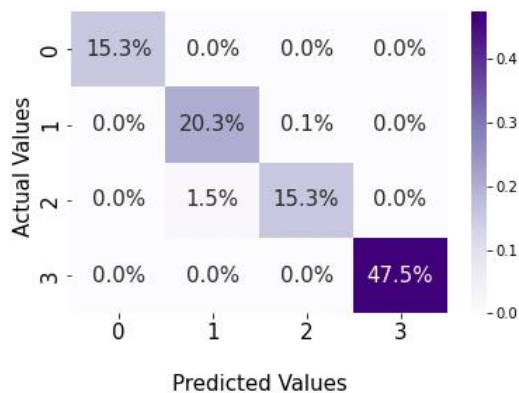

**FIGURE 5: Clustering accuracy of the test set**

The MC task accuracy is 98.4%, indicating the acceptable performance of this method, which proves the efficiency of the Silhouette analysis in determining the optimal number of clusters and verifies the performance of the DNN clustering technique. This should be highlighted that with incorrect clustering, the micro-clustering impact will vanish, and we will lose the benefits of the micro-clustering task. In this case, the accuracy of the predicted outputs will be worse than with approaches without a clustering task, which is indicated in the numerical results as ANN or DNN. Micro clustering is the core of the proposed methodology, and it is strongly recommended to validate the classification task's accuracy before implementing it in the proposed method.

B.  Forecasting Result:

**Table 4** shows the estimated error criteria measures for several forecasting algorithms, including the proposed MC-DNN methods. According to the results, the proposed method has the best rate in four of the five error criteria measures. Accordingly, it is possible to conclude that the suggested method outperforms previous benchmark methods in predicting charging behavior. **Table 4** includes recall, precision, F-measures, and G-Mean, which are the most significant error criteria for imbalanced classification issues. Consequently, these error criteria provide a better knowledge of the performance of various methods.

**TABLE 4: Accuracy measures result**

| Methods | Accuracy | Precision | Recall | F-Measure | G-Mean |
|---------|----------|-----------|--------|-----------|--------|
| SVM-Poly | 0.645 | 0.476 | 0.755 | 0.514 | 0.599 |
| SVM-RBF | 0.702 | 0.503 | 0.734 | 0.553 | 0.607 |
| KNN | 0.712 | 0.494 | 0.644 | 0.54 | 0.561 |
| CS-LR | 0.751 | 0.542 | 0.669 | 0.579 | 0.602 |
| ANN | 0.743 | 0.567 | 0.753 | 0.614 | 0.653 |
| DNN | 0.761 | 0.578 | 0.750 | 0.636 | 0.658 |
| **MC-DNN** | **0.809** | **0.652** | **0.725** | **0.684** | **0.687** |

To compare different methods, **Figure 6** visualizes the error criteria for different measures.

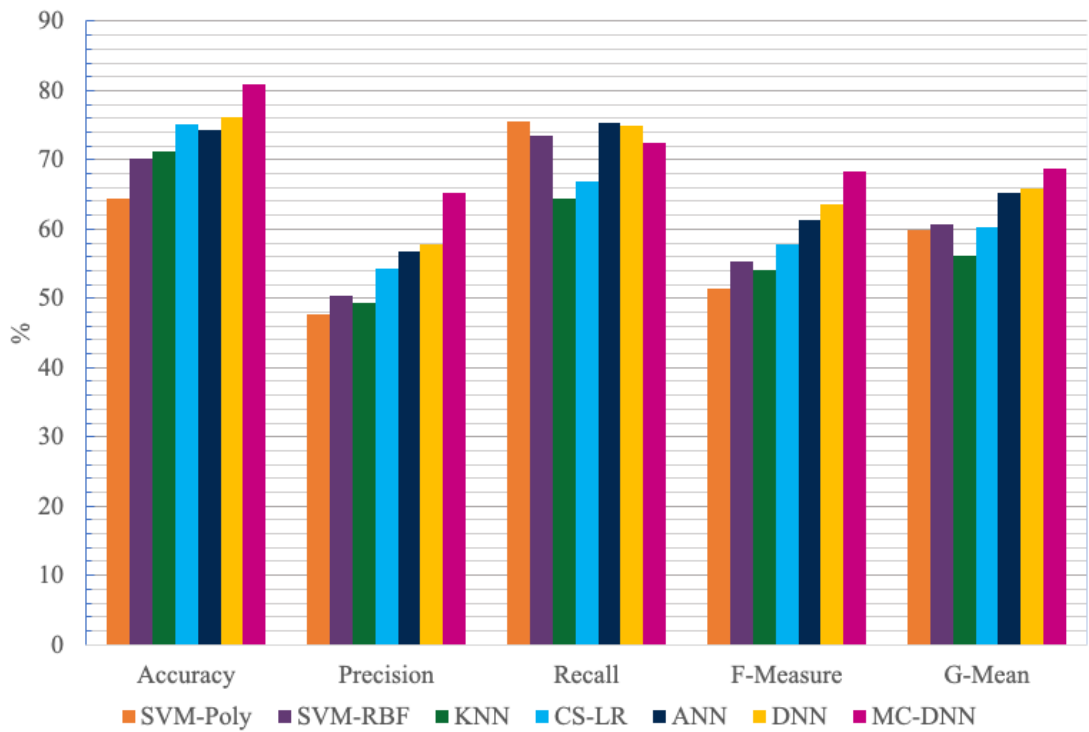

**FIGURE 6: Accuracy Measures for Considered Models**

## IV. DISCUSSION

To verify the robustness of the proposed method, the dataset was split into training, validation, and test sets – 60%, 20%, and 20%, respectively – and a full comparison of various approaches was considered. **Figure 7** depicts the percentage of correctly predicted Level 1, Level 2, and DC-Fast charging events. Also, it shows the percentage of events that were underestimated and classified as the wrong class. According to the findings, both SVM algorithms have the lowest percentage of incorrect predictions in Level 1 and DC-Fast charging. However, for Level 2 which has the highest number of samples among the three charging levels, the MC-DNN has the second-highest performance after the DNN (**Figure 7-b**).

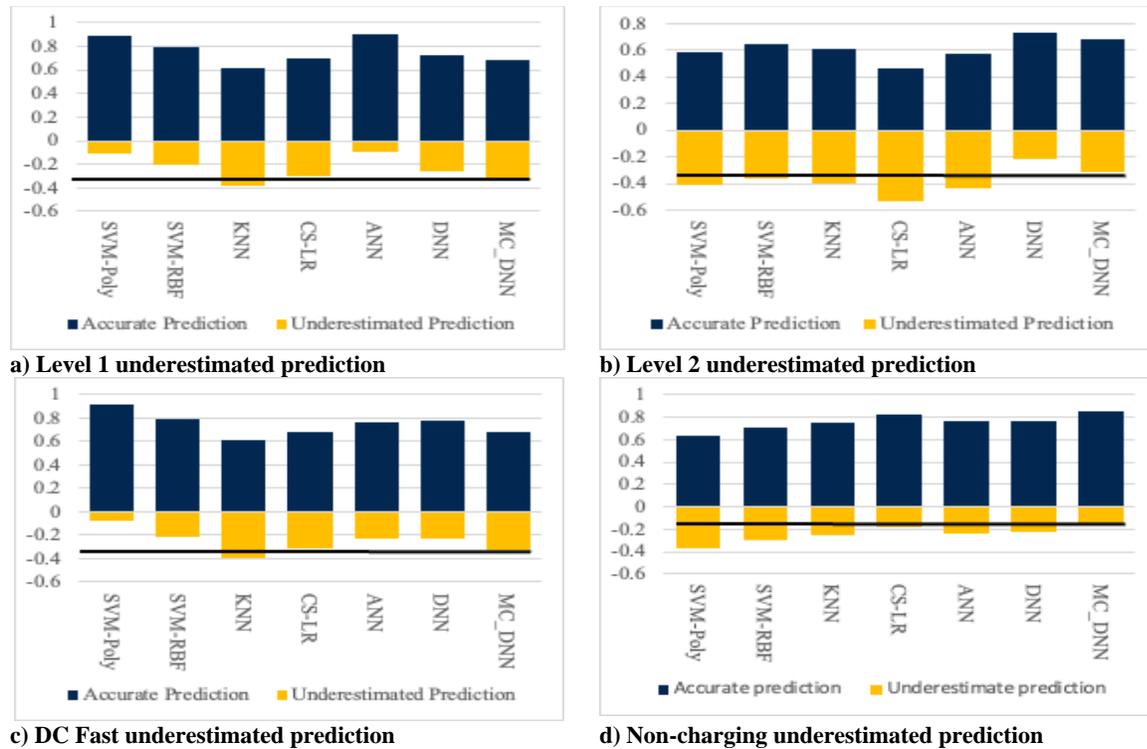

a) Level 1 underestimated prediction    b) Level 2 underestimated prediction

c) DC Fast underestimated prediction    d) Non-charging underestimated prediction

**FIGURE7: Underestimated Charging Predictions**

However, all methods should be examined to determine which can reduce overestimated charging events. Because of the imbalanced classes, some of the samples in the largest class, which is non-charging event in this study, could be predicted as one of these three charging classes. Thus, it is vital to determine which approach has the highest performance to reduce the number of incorrectly predicted events. **Figure 8** shows that the proposed method performs better than all other methods and is able to reduce the number of wrongly predicted events in all three charging classes. Because the outcome of these studies is crucial for building charging stations and convincing decision-makers to invest in infrastructure, even a 1% increase in accuracy could reduce the amount of investment. Particularly for the most expensive one, which is the DC-Fast charging station. Each DC Fast charger installation costs between $20,000 and $150,000 (*38, 39*).

As a result, charging event estimation is crucial to the number of charging events for the BEV fleet and the number of charging stations. The overestimated prediction rate is more critical than the underestimated prediction rate from an investment perspective. Because the investment will be made based on the number of charging events predicted in each class. The lower number of wrong predictions means decreased investment in unnecessary charging stations. The less overestimated event is more critical when the station installation cost increases, especially for DC-Fast stations. The proposed method shows improvement in dropping the wrong charging event prediction, which helps charge providers avoid financial loss. Based on the results, the MC-DNN method has the highest accuracy and F-measure in charging event prediction, followed by the DNN and ANN.

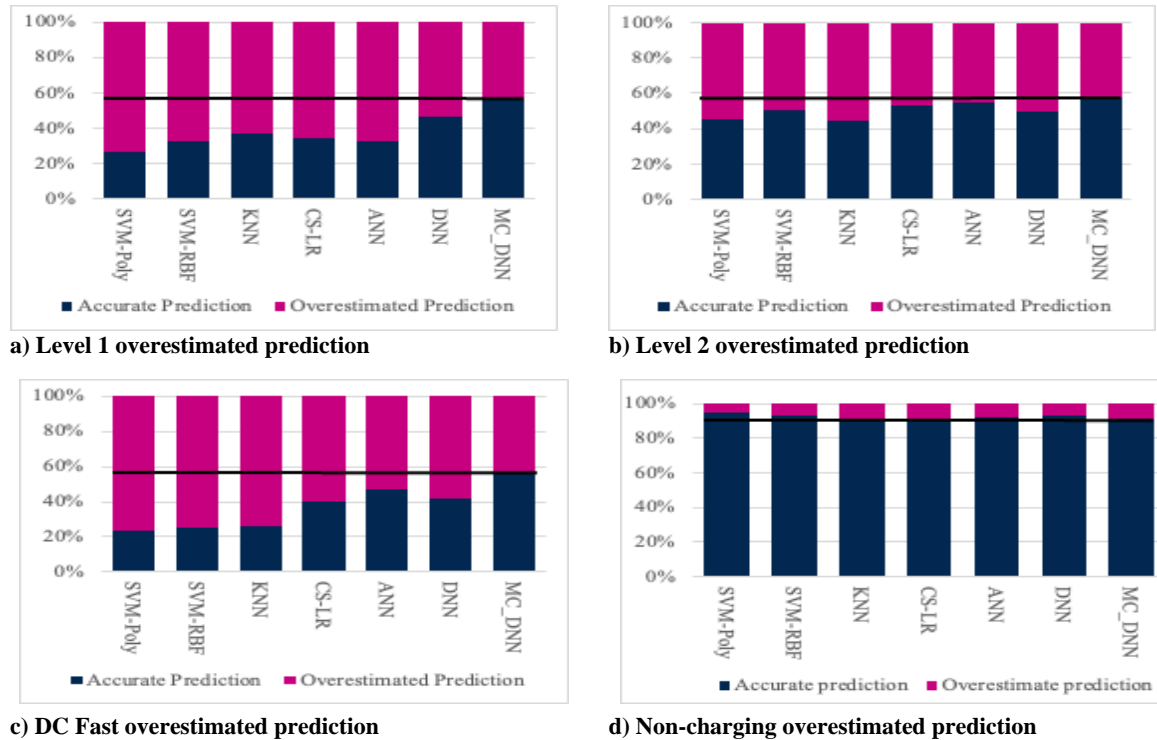

a) Level 1 overestimated prediction        b) Level 2 overestimated prediction

c) DC Fast overestimated prediction        d) Non-charging overestimated prediction

**FIGURE 8: Overestimated Charging Predictions**

One of the largest DC-Fast charging network infrastructures in the US, Electrify America, recently shared information about how the stations were used in 2021. Their report (*40*) stated that, on average, five charging events happened on DC-Fast charging stations daily on fast-charging stations, which is equal to 35 charging events per week on average. Thus, reducing the number of overestimated charging events as much as possible could help the operator decrease the investment and optimize the charging stations' utility factor. For instance, by utilizing MC-DNN, the number of overestimated DC-Fast charging events decreased by 9.17 percent compared to DNN, which is the second most accurate technique, as shown in **Table 4**.

## V. CONCLUSION

The electrification of transportation is being pushed with the expectation of lowering emissions and enhancing public health. While the prototype results satisfied the consequences of significant BEV market adoption on charging infrastructure, the future of charging infrastructure remains unknown. To properly comprehend the charging infrastructure needs of BEVs at high market penetration, the stochastic charging behavior of BEV users must be predicted. Because of their driving patterns, BEV customers have quite complex charging behavior. As a result, the need for a prediction approach that can handle large amounts of data for real-world applications while assuming high accuracy has become apparent. Indeed, robust forecasting tools based on artificial intelligence are required to model high-dynamic behavior profiles such as charging event behavior. Modeling charging behavior is significant since BEV charging events have an impact not just on transportation systems but also on the energy sector. This study utilizes a micro-clustering technique and a DNN method to provide a more accurate method for predicting charging events, which is necessary for estimating the number of charging stations in the future. The charging level is the most important aspect of the charging behavior of BEV users since it affects their energy consumption and charging patterns. The proposed micro-clustering DNN model, which is developed by coupling unsupervised and supervised clustering tasks, exceeded existing benchmark methods in terms of output quality. The model achieves 80.9% accuracy in predicting charging events and 68.7% F-Measures. The numerical results of this study show the performance of the MC-DNN in forecasting charging events. For instance, the number of overestimated DC-Fast charging events decreased by 9.17% utilizing the MC-DNN method.

To extract the embedded activity information, our model could be used to predict the charging behavior at the beginning of a trip for every individual activity trace, including new real BEV information. Besides introducing a novel method for predicting charging events, the significant contribution of this paper is using actual BEV travel and charging patterns to train a model and using specific charging levels for each charging event, since other studies usually ignore the charging level.

Lastly, for future work, Generative Adversarial Networks (GAN) approaches could be used along with our model to map the current internal combustion engine (ICE) vehicles' information and create a synthetic dataset based on their behavior. This would allow us to estimate and assess changes in investment in charging infrastructure under future scenarios in which the market share of the BEV fleet grows. This would help develop charging infrastructure and pave the way for a larger BEV market share.

## ACKNOWLEDGMENTS

The authors would like to thank the California Air Resource Board and the Institute of Transportation Studies at the University of California, Davis for funding this project. The authors would like to thank the researchers and students of the Plug-in Electric Vehicle Research Center at the University of California who contributed to this study. The data analyzed during this study is not available to the public.

## AUTHOR CONTRIBUTIONS

The authors confirm their contribution to the paper as follows: Study conception and design: HT, TR, VK, GT, CN; Data collection: HT, VK, GT, CN; Analysis and interpretation of results: HT, TR, CN, GT; Draft manuscript preparation: HT, TR, GT, CN. All authors reviewed the results and approved the final version of the manuscript.